\documentclass[]{gtech}

\usepackage{amssymb}
\usepackage{multirow}
\usepackage{bigdelim}
\usepackage{todonotes}
\usepackage{longtable}
\usepackage{tabularray}
\usepackage{wrapfig}
\usepackage[most]{tcolorbox} 

\hypersetup{
  colorlinks=true,
  citecolor=green!50!black,
  linkcolor=red!60!black,
  urlcolor=green!50!black
}
\usepackage{xcolor}
\usepackage{url}
\usepackage{mathtools}
\usepackage{amsmath}
\usepackage{amsfonts}
\usepackage{booktabs}

\usepackage{orcidlink}
\usepackage{nicefrac}
\usepackage{microtype}
\usepackage{algorithm}
\usepackage{algpseudocode}
\usepackage{CJK}
\usepackage{subcaption}
\usepackage{cleveref}
\usepackage{enumitem}
\usepackage{mathrsfs}
\usepackage{amsthm}
\usepackage{tikz}
\usepackage[table]{xcolor}

\renewcommand{\title}[1]{\newcommand{\titlelist}{{\huge\fontfamily{optimistic}\selectfont #1}}}

\definecolor{tcaeblue}{HTML}{0369FF}
\newcommand{\model}{\texttt{\textbf{\textcolor{tcaeblue}{TC-AE}}}}
\newcommand{\cmark}{\ding{51}}%
\newcommand{\xmark}{\ding{55}}%

\definecolor{podiblue}{rgb}{0.9, 0.92, 1.0}
\definecolor{rowblue}{RGB}{200, 225, 245}
\definecolor{deepred}{RGB}{235, 120, 120}
\definecolor{prompt}{HTML}{5f84e4}
\definecolor{img}{HTML}{820100}

\usepackage{arydshln}

\usepackage{colortbl}
\usepackage{amssymb}
\usepackage{pifont}
\usepackage{booktabs,multirow}
\usepackage{makecell}
\usepackage{tabulary}
\usepackage{fontawesome5}
\usepackage{bbding}
\usepackage{multicol}

\newcommand{\finding}[2]{
    \vspace{-0.1cm}
    \begin{tcolorbox}[
        colback=white!90!gray,
        colframe=teal!60!black,
        arc=5pt,
        boxsep=5pt,
        left=6pt,
        right=6pt,
        top=2pt,
        bottom=2pt,
        boxrule=0.8pt,
        drop shadow=gray!50!white,
        enhanced jigsaw
    ]
    \vspace{-0.1cm}
        \paragraph{\textbf{\textit{Observation #1:}}}#2
    \vspace{-0.1cm}
    \end{tcolorbox}
    \vspace{-0.1cm}
}



\newlength\savewidth

\def\github{\raisebox{-1.5pt}{\includegraphics[height=1.05em]{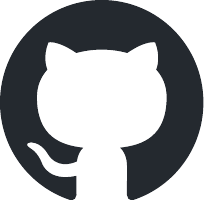}}}

\title{\textcolor{tcaeblue}{TC-AE}: Unlocking Token Capacity for Deep Compression Autoencoders}

\author[1,2,*]{Teng Li}
\author[1,*,\dagger]{Ziyuan Huang}
\author[1,3,*]{Cong Chen}
\author[1,4]{Yangfu Li}
\author[1,5]{Yuanhuiyi Lyu}
\author[]{}

\secondauthor[1]{Dandan Zheng}
\secondauthor[3]{Chunhua Shen}
\secondauthor[2,\dagger]{Jun Zhang}

\affiliation[1]{Inclusion AI, Ant Group}
\affiliation[2]{HKUST}
\affiliation[3]{ZJU}
\affiliation[4]{ECNU}
\affiliation[5]{HKUST (GZ)}

\contribution[*]{Equal contribution}
\contribution[\dagger]{Corresponding authors}

\abstract{\fontsize{11pt}{12pt} 
\textit{We propose \model{}, a Vision Transformer (ViT)– based architecture for deep compression autoencoders. Existing methods commonly increase the channel number of latent representations to maintain reconstruction quality under high compression ratios. However, this strategy often leads to latent representation collapse, which degrades generative performance. Instead of relying on increasingly complex architectures or multi-stage training schemes, TC-AE addresses this challenge from the perspective of the token space, the key bridge between pixels and image latents, through two complementary innovations: (1) We study token number scaling by adjusting the patch size in ViT under a fixed latent budget, and identify aggressive token-to-latent compression as the key factor that limits effective scaling. To address this issue, we decompose token-to-latent compression into two stages, reducing structural information loss and enabling effective token number scaling for generation. (2) To further mitigate latent representation collapse, we enhance the semantic structure of image tokens via joint self-supervised training, leading to more generative-friendly latents. With these designs, TC-AE achieves substantially improved reconstruction and generative performance under deep compression. We hope our research will advance ViT-based tokenizer for visual generation.
}}

\date{April 9, 2026}
\gtechdata[\github]{\url{https://github.com/inclusionAI/TC-AE}} 
\begin{document}
\maketitle

\section{Introduction}

Latent diffusion models (LDMs)~\citep{ldm} have emerged as the dominant paradigm for efficient image generation. They typically adopt a two-stage framework: a tokenizer compresses images into latent representations and a diffusion model operates in the latent space.
To further improve efficiency, recent works~\citep{dc-ae, ming-univision} increasingly push tokenizers toward deep compression by aggressively reducing the spatial resolution of the latent representation, while compensating with a larger channel number to preserve reconstruction quality.
However, this combination of extreme spatial compression and channel expansion often leads to latent representation collapse, resulting in degraded generative performance~\citep{dc-ae-1.5}.
In this work, we study this challenge in the context of ViT-based tokenizers, due to their strong potential in scalability and representation capacity~\citep{vitok, gigatok}.

\begin{figure*}[t]
    \centering
    \includegraphics[width=0.95\linewidth]{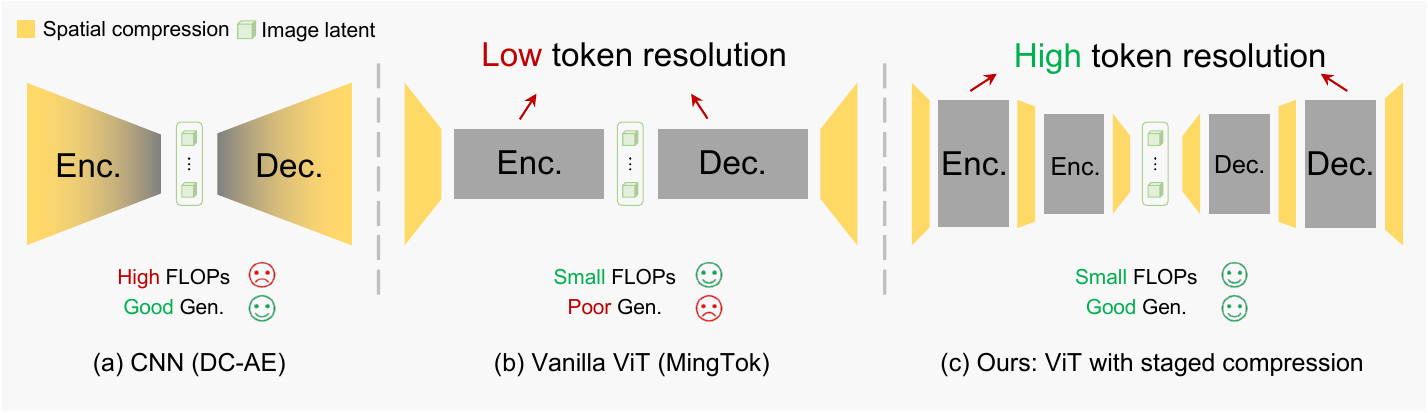}
\caption{Architectural Comparasion of existing methods. (a) CNN-based tokenizers achieve strong generation with high FLOPs. (b) Vanilla ViT-based tokenizers are computationally efficient but suffer from severe information loss in the patchify layer, as spatial resolution is not gradually reduced as in CNNs. (c) TC-AE introduces staged token compression in the ViT encoder, balancing generative performance and efficiency.}
    \label{fig:arch_comparison}
\end{figure*}

In ViT-based tokenizers~\citep{vitok, hieratok, ming-univision}, images are firstly patchified into a sequence of image tokens, which are then processed by Transformer blocks, and finally compressed into compact latents through a bottleneck. Image tokens therefore serve as a critical bridge between raw pixels and the latent space, and their capacity and structure directly influence the quality of the resulting latent representations. 
Existing approaches primarily focus on scaling model parameters~\citep{vitok, gigatok} or directly optimizing the final latent representations~\citep{vavae, ming-univision} to improve generative performance, while overlooking the optimization of the intermediate token space. Meanwhile, studies on ViT-based representation learning~\citep{dino, patch-scaling} have shown that increasing the number of tokens can substantially improve downstream performance and lead to more semantically structured representations. However, whether and how scaling the token space translates into improved latent generatability in LDM frameworks remains an open problem.

\begin{figure*}[t]
    \centering
    \includegraphics[width=0.95\linewidth]{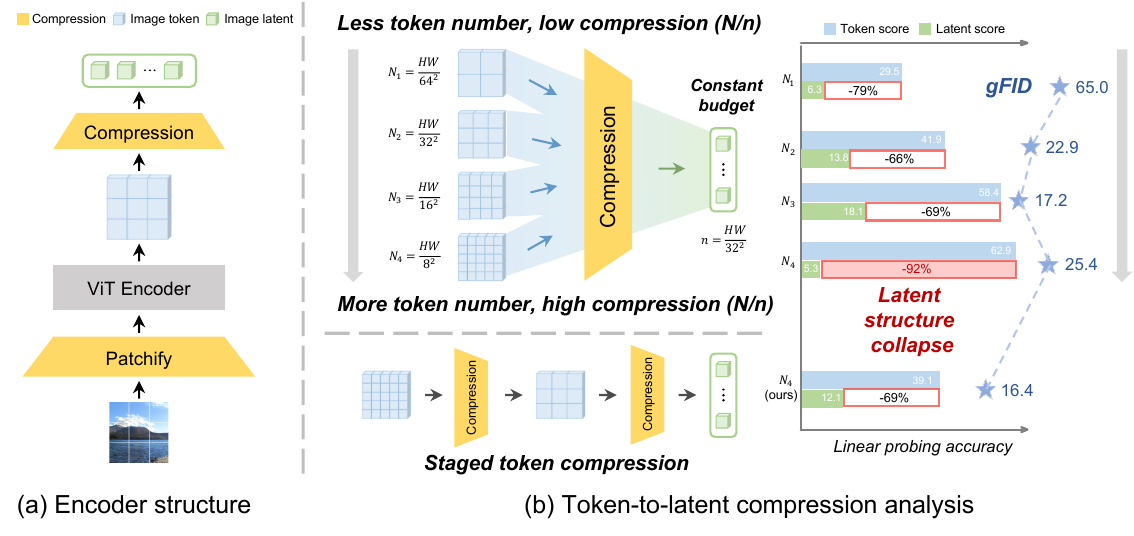}
\caption{Latent structure collapse caused by token-to-latent compression when scaling token numbers. Although increasing token numbers improves the token-level structure, the fixed latent resolution forces more aggressive token-to-latent compression, causing severe information loss and latent structure collapse at small patch sizes (\textit{e.g.}, $p=8$).}
    \label{fig:bottleneck}
\end{figure*}

In our preliminary experiments, we observe that naively increasing the number of image tokens consistently improves reconstruction quality, but does not necessarily lead to better generation performance (Fig.~\ref{fig:scaling_pz}). We further identify the token-to-latent compression as the root cause of this phenomenon (Fig.~\ref{fig:bottleneck} and Sec.~\ref{sec: staged compression}).
Under a fixed pixel-to-latent compression ratio, increasing the intermediate token number inevitably amplifies the compression ratio at the token-to-latent compression bottleneck. Such aggressive compression causes substantial structural information loss in the latent space, which in turn degrades generative performance.
To address this issue, we propose staged token compression, which decomposes the token-to-latent bottleneck into two stages: an intermediate compression stage applied within the ViT encoder to reduce token resolution, followed by a final bottleneck that further compresses the tokens into compact image latent.
Unlike vanilla ViT-based tokenizers that rely on a single abrupt bottleneck, this design preserves latent structure under large token spaces and restores positive scaling trends for both reconstruction and generation.

Beyond increasing token capacity, we further investigate improving the semantic structure of the token space to enhance latent generatability.
Recent studies~\citep{vavae, gigatok, ming-univision} have shown that semantically regularized latent representations can significantly benefit generative modeling, often achieved by aligning latents with representations from pretrained foundation models~\citep{dinov2, mae}.
However, such approaches rely on external large-scale pretraining and implicitly introduce strong data priors.
In contrast, we introduce a joint self-supervised objective directly into tokenizer training, enabling the ViT encoder to produce semantically structured image tokens.
This design consistently improves generation performance across different token numbers.

Coupling these designs, we propose \model{}, a ViT-based tokenizer powered by a large and semantically structured token space for image generation.
Fig.~\ref{fig:arch_comparison} presents a high-level framework comparison between TC-AE and other deep compression tokenizers. Fig.~\ref{fig:exp:scaling} shows that token number scaling and model parameter scaling are complementary, jointly improving generative performance. {\color{tcaeblue}\textbf{\textit{This highlights token space optimization as a new and effective direction for ViT-based tokenizers.}}} Our main contributions are summarized as follows:

\begin{enumerate}[label=\textbullet, leftmargin=10pt, itemsep=3.5pt, topsep=2pt]

\item \textbf{Novel perspective.}
We are the first to address representation collapse in deep compression autoencoders from the perspective of the token space, and identify the token-to-latent compression as a key factor that limits effective token number scaling.
\item \model{}.
We introduce staged token compression to alleviate structural information loss during token-to-latent mapping, and further improve token structure via joint self-supervised training.
\item \textbf{Strong performance.}
TC-AE achieves consistently better reconstruction and generative performance than existing tokenizers, and shows that token number scaling offers a more effective and complementary scaling dimension than parameter scaling.

\end{enumerate}

\section{Related Work}
\subsection{Deep Compression Visual Tokenizers}
The primary goal of visual tokenizers is to compress images into compact latent representations for efficient generative modeling~\citep{ldm}. Recently, achieving extremely high compression ratios~\citep{dc-ae, dc-ae-1.5, ming-univision} has become an active research focus to further improve generation efficiency~\citep{dc-gen, sana-1.5}. However, such deep compression typically reduces the spatial resolution of the latent representation, often leading to severe reconstruction degradation and representation collapse. To alleviate these issues, prior works~\citep{dc-ae, dc-ae-1.5} commonly increase the latent channel dimension to compensate for the loss of spatial capacity, and employ multi-stage training strategies to encourage more structured latent representations.
While effective to some extent, these approaches introduce additional architectural and training complexity. In contrast, our work addresses the challenges of deep compression from a different perspective: scaling the token space. We investigate how both reconstruction and generation performance scale with the number of image tokens, and propose a staged token compression architecture that mitigates the trade-off between compression efficiency and representation quality.

\subsection{Scaling Token Space in ViT}
In ViT-based tokenizers~\citep{vitok, ming-univision, hieratok}, the image token acts as a critical bridge connecting raw RGB inputs to the latent space. Increasing the number of tokens theoretically provides a more comprehensive representation of the visual content. 
This benefit of increasing token granularity has been widely observed in visual representation learning. In self-supervised ViTs, increasing the number of tokens leads to richer semantic representations and improved transfer performance, as demonstrated by DINO~\citep{dino} and subsequent patch-level representation analysis~\citep{patch-representation}. More recent work further establishes systematic scaling laws for patchification, showing consistent gains across downstream tasks as token counts increase~\citep{patch-scaling}. Similar trends have also been reported in generative modeling~\citep{dit}, where smaller patch size yields better generation performance.
However, we are the first to systematically introduce this token scaling perspective into the design of tokenizers specifically for generation. Our investigation reveals a critical trade-off between reconstruction and generation in naive scaling strategies, and we propose a staged token compression strategy that resolves this conflict.

\subsection{Representation Learning in Generation}
The integration of representation learning into generative frameworks has recently achieved remarkable success. For example, REPA~\citep{repa} and iREPA~\citep{irepa} improve generative performance by aligning intermediate representations in the SiT~\citep{sit} framework with features from DINOv2~\citep{dinov2}. Similarly, in the context of tokenizer design, approaches such as VA-VAE~\citep{vavae}, MAETok~\citep{maetok}, and GigaTok~\citep{gigatok} incorporate pretrained visual foundation models to guide latent optimization, while RAE~\citep{rae} directly adopts representations from foundation models for generation.
Despite their effectiveness, these strategies typically rely on heavy teacher models and implicitly introduce strong priors from large-scale pretraining. In contrast, we adopt a self-supervised joint training mechanism that structures the token space during tokenizer training, enabling us to demonstrate the intrinsic value of representation learning for deep compression tokenizers.

\section{Method}
\subsection{Preliminary: Compression in ViT-based Image Autoencoder}

Recent image tokenizers~\citep{vitok, gigatok, hieratok, vtp} increasingly adopt the Vision Transformer (ViT)~\citep{vit} architecture due to its strong scalability and representation capacity.
A vanilla ViT-based image autoencoder consists of an encoder $\mathcal{E}$ and a decoder $\mathcal{D}$, which compress an input image into a latent representation and reconstruct it back to the pixel space.

Given an input image $\mathbf{X} \in \mathbb{R}^{H \times W \times 3}$, the encoder applies a patch embedding layer $\phi_p(\cdot)$ that partitions the image into a 2D grid of non-overlapping patches with size $p\times p$, projects each patch to a $d$-dimensional embedding, and flattens the grid into a token sequence:

\begin{equation}
\mathbf{T} = \phi_{p}(\mathbf{X}) \in \mathbb{R}^{N \times d}, \qquad 
N = \frac{H \cdot W}{p^2}.
\end{equation}

\begin{figure*}[t]
    \centering
    \includegraphics[width=\linewidth]{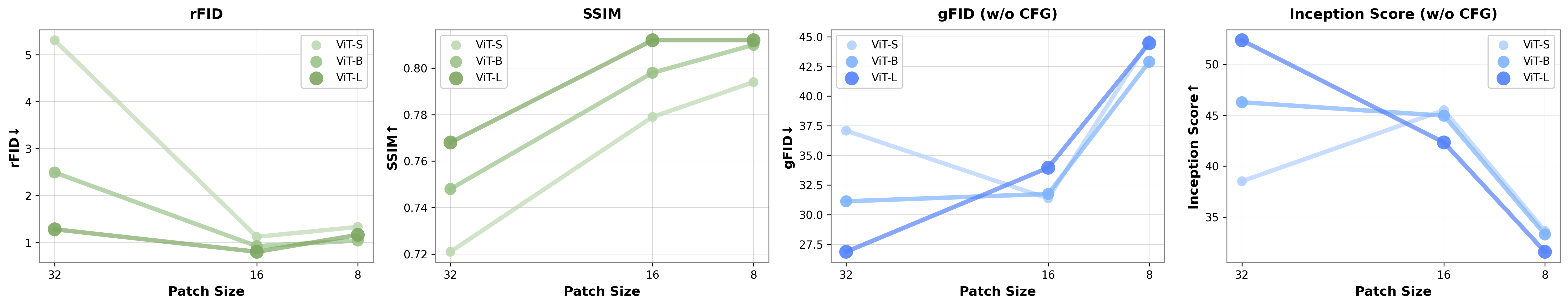}
\caption{Naive scaling of token numbers for reconstruction and generation. We increase the number of image tokens by reducing the ViT patch size. While reconstruction quality consistently improves with more tokens (shown in the left two figures), generative performance does not exhibit corresponding gains (shown in the right two figures).}
    \label{fig:scaling_pz}
\end{figure*}

The image tokens $\mathbf{T}$ are processed by a stack of Transformer layers $\mathrm{TF}(\cdot)$ and subsequently compressed by a bottleneck layer $\mathcal{B}(\cdot)$ into a latent tensor:

\begin{equation}
\mathbf{z} = \mathcal{B}\!\left(\mathrm{TF}(\mathbf{T})\right) \in \mathbb{R}^{h \times w \times c},
\end{equation}
where $(h, w)$ denotes the spatial resolution of the latent representation $\mathbf{z}$ and $c$ is the channel dimension.
The latent $\mathbf{z}$ then serves as the learning objective for downstream generative models.

\noindent \textbf{Image tokens as the information bridge.}
In a ViT-based autoencoder, the information flow from the input image to the latent representation involves two consecutive compression stages: pixel-to-token compression in the patch embedding layer, and token-to-latent compression at the bottleneck.
The overall spatial compression ratio $f_{\text{pix}\rightarrow\text{lat}}$ can therefore be decomposed as:

\begin{equation}
\label{eq:compression_ratio}
f_{\text{pix}\rightarrow\text{lat}}
= f_{\text{pix}\rightarrow\text{tok}} \times f_{\text{tok}\rightarrow\text{lat}}
= p^2 \cdot \frac{N}{h \cdot w},
\end{equation}
where $f_{\text{pix}\rightarrow\text{tok}}$ and $f_{\text{tok}\rightarrow\text{lat}}$ denote the spatial compression introduced at the patch embedding and bottleneck stages, respectively.

Image tokens therefore act as an information bridge between the input image and the latent space.
Their capacity and semantic structure directly determine how effectively information is preserved through compression.
In the following sections, we study how improving the token space impacts the quality of the latent representation and, consequently, the reconstruction and generative performance.

\subsection{Naive Scaling Token Numbers Fails to Improve Generation}

Increasing the number of image tokens in ViT has been shown to improve model capacity~\citep{dino,patch-scaling}. Motivated by this observation, we investigate whether token number scaling can similarly benefit deep compression autoencoders.

We conduct controlled experiments under a fixed latent budget. Images are spatially compressed by a factor of $f_{\text{pix}\rightarrow\text{lat}} = 32$, and the latent channel dimension is fixed to $c = 128$ across all settings, following common practice in deep compression autoencoders~\citep{dc-ae-1.5}.
The number of image tokens is increased by sweeping the patch size $p \in \{32,16,8\}$ for different tokenizer model sizes (ViT-S, ViT-B, and ViT-L). As shown in Fig.~\ref{fig:scaling_pz}, reconstruction quality consistently improves as the number of image tokens increases, indicating that higher token resolution enables the tokenizer to capture finer-grained visual details.
In contrast, generation performance does not improve with token number scaling. 
While similar limitations have been reported in prior works when scaling model parameters~\citep{vitok, gigatok}, our results reveal that this issue persists when scaling along the token space dimension via patch size reduction.

\finding{1}{Scaling the number of image tokens improves reconstruction quality, but it does not improve generative performance under a fixed latent budget.}

\subsection{Scaling Token Numbers with Staged Token Compression}
\label{sec: staged compression}

Prior works~\citep{vavae, gigatok, rae, svg-tokenizer} have shown that enriching latent representations with semantic structure improves their generative quality.
Motivated by this observation, we analyze the failure of token number scaling from the perspective of semantic preservation.

Specifically, we examine how semantic information carried by image tokens is retained when compressed into the latent representation using linear probing~\citep{gigatok}.
We train linear classifiers on spatially average-pooled image tokens (before bottleneck) and latent features (after bottleneck), and denote the resulting classification accuracies as $A_1$ and $A_2$, respectively.
The ratio $A_2/A_1$ serves as a quantitative measure of semantic information preserved across the compression bottleneck $\mathcal{B}(\cdot)$, while $(1-\frac{A_2}{A_1})$ represents the structure loss in this process.

\begin{figure*}[t]
    \centering
    \includegraphics[width=\linewidth]{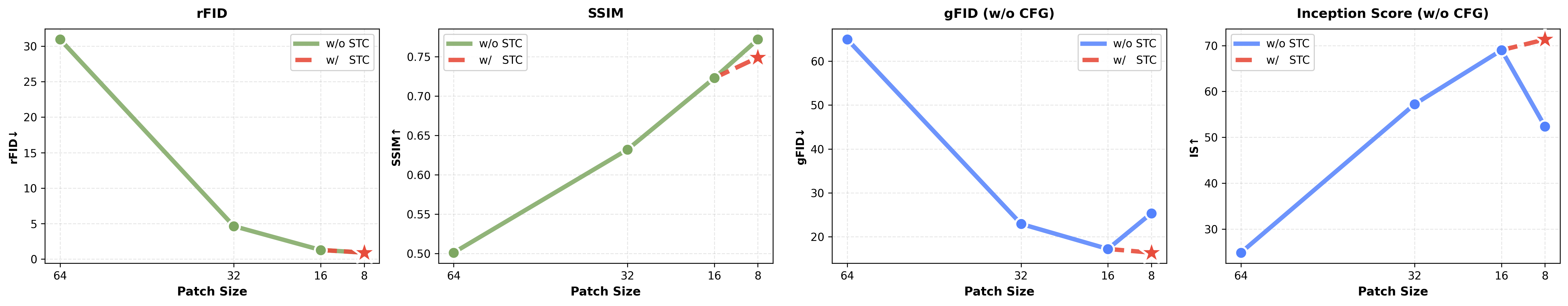}
\caption{Token number scaling with staged token compression (STC).
This strategy mitigates structural information loss at the bottleneck under large token spaces, enabling generative performance to scale with increasing token numbers.}
    \label{fig:scaling_pz_ssl_compress}
\end{figure*}

\noindent \textbf{Bottleneck-induced latent structure loss.}
Results are reported in Table~\ref{tab:linear_probing}.
As the patch size decreases, the semantic structure of image tokens improves as indicated by increasing $A_1$.
However, this improvement does not carry over to the latent space: at smaller patch sizes (e.g., $p=8$), $A_2$ drops sharply, resulting in a very high information loss ($1-\frac{A_2}{A_1} = 0.92$). This indicates severe semantic loss during token-to-latent compression.

This structure loss can be explained by the compression budget constraint in Eq.~\ref{eq:compression_ratio}.
Under a fixed pixel-to-latent compression ratio $f_{\text{pix}\rightarrow\text{lat}}$, decreasing the patch size reduces the pixel-to-token compression factor $f_{\text{pix}\rightarrow\text{tok}}$, but necessarily increases the token-to-latent compression factor $f_{\text{tok}\rightarrow\text{lat}}$.
Consequently, although image tokens become increasingly semantically structured, the aggressive compression at the bottleneck destroys this structure, preventing generative performance from scaling with the number of tokens.

\begin{table}[tbp]
  \centering
    \caption{Latent structure loss under token-to-latent compression.
    Increasing token numbers improves semantics of image tokens ($A_1$), but aggressive bottleneck compression causes severe semantic loss in the latent space.
    Bottl.: bottleneck. Str.: structure.}
      \resizebox{0.77\linewidth}{!}{ 
    \begin{tabular}{ccccc}
    \toprule
    \multirow{2}[2]{*}{Patch Size} & \multirow{2}[2]{*}{gFID$\downarrow$} & \multicolumn{2}{c}{Linear Probing Accuracy} \\
          &       & Before Bottl. ($A_1$)$\uparrow$ & After Bottl. ($A_2$)$\uparrow$ & Str. Loss ($1-\frac{A_2}{A_1}$)$\uparrow$ \\
    \midrule
    64    & 64.96 & 29.5  & 6.3   & 0.79 \\
    32    & 22.93 & 41.9  & 13.8  & 0.67 \\
    16    & 17.22 & 58.4  & 18.1  & 0.69 \\
    8     & 25.36 & 62.9  & 5.33  & \cellcolor{deepred}{0.92}\\
    \midrule 
    8 (w/ STC)     & 16.39 & 39.1  & 12.1  & \cellcolor{rowblue}{0.69} \\
    \bottomrule
    \end{tabular}}%
  \label{tab:linear_probing}%
\end{table}%

\noindent \textbf{Staged token compression.}
To address this issue, we propose staged token compression to redistribute the compression process from the bottleneck into the ViT Transformer layers. Specifically, we divide the layers into two stages. The first stage operates on a large number of tokens to learn rich semantic structure, followed by an intermediate compression layer that aggregates semantics into a small set of tokens. Then, a second compression is applied at the bottleneck to produce the final latent representation. By decomposing a single aggressive compression step into multiple steps, image latent preserves semantic structure more effectively during token-to-latent transformation.

As shown in Table~\ref{tab:linear_probing}, staged token compression substantially increases the semantic preservation ratio $A_2/A_1$ from $0.08$ to $0.31$.
Consequently, as shown in Fig.~\ref{fig:scaling_pz_ssl_compress}, it achieves the strongest generative performance and enables generation quality to continue improving with increasing numbers of image tokens.

\finding{2}{We identify the token-to-latent compression bottleneck as the primary factor limiting token number scaling, and it can be addressed by with staged token compression.}

\subsection{Enhancing Token Structure via Self-Supervision}

Beyond increasing the number of tokens, we investigate enhancing the semantic structure of the token space to improve the generative quality of the latent.
Prior works~\citep{vavae, gigatok, rae, svg-tokenizer} typically achieve this by aligning latent with pretrained models or initializating encoder from their weights.
However, such approaches rely on external large-scale pretraining and require the pretraining configuration (e.g., image resolution and compression ratio) to be tightly coupled with the tokenizer design.
In contrast, we introduce joint self-supervised learning (SSL) as an auxiliary objective to structure the token space during tokenizer training.

\noindent \textbf{Self-supervised auxiliary training with iBOT.}
We incorporate iBOT~\citep{ibot} as an auxiliary training objective. iBOT employs a student--teacher distillation framework. The teacher is an exponential moving average (EMA) of the student.

Given an input image $\mathbf{X}$, teacher and student inputs are generated via two augmentation pipelines, $Aug_{tea}(\cdot)$ and $Aug_{stu}(\cdot)$.
The teacher pipeline $Aug_{tea}(\mathbf{X})$ produces two global crops.
The student pipeline $Aug_{stu}(\mathbf{X})$ produces the same global crops with random patch masking, along with multiple additional local crops.
For the masked global crops, the student is trained to predict the teacher patch-token outputs, corresponding to a masked image modeling objective $\mathcal{L}_{\text{MIM}}$.
For the local crops, the student class-token predictions are aligned with those of the teacher to enforce semantic consistency across different views, yielding the class-token distillation loss $\mathcal{L}_{[\text{CLS}]}$.
Together, these objectives encourage both local-level and global-level semantic structure. The overall self-supervised objective is given by:
\begin{equation}
\mathcal{L}_{\text{iBOT}} = \mathcal{L}_{\text{MIM}} + \mathcal{L}_{[\text{CLS}]}.
\end{equation}

\begin{figure*}[t]
    \centering
    \includegraphics[width=0.98\linewidth]{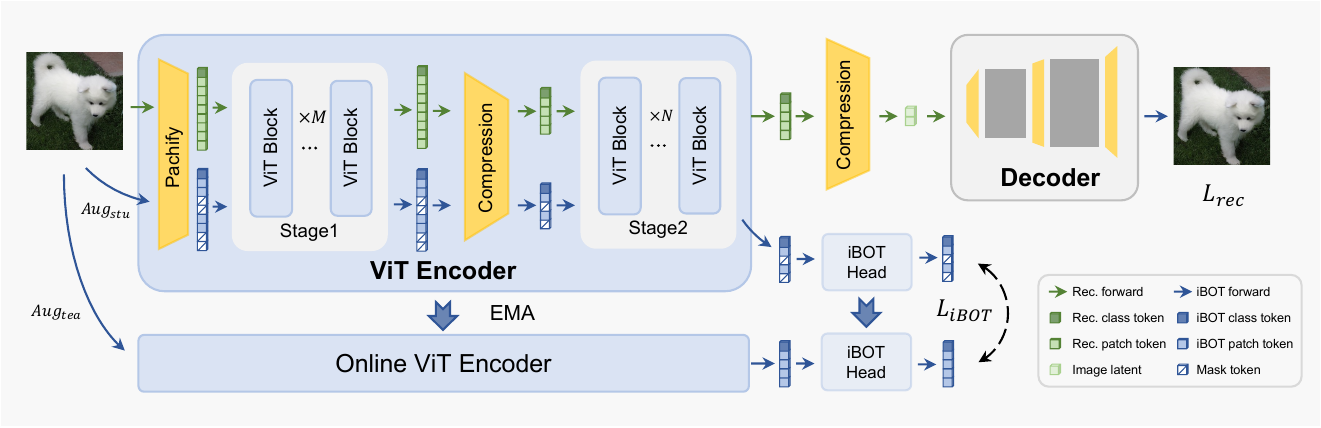}
\caption{Architecture of TC-AE.
TC-AE introduces staged token compression to alleviate structural information loss at the bottleneck, enabling generative performance to scale with the number of image tokens.
In addition, self-supervised training is incorporated to further structure the token space.}
    \label{fig:main_arch}
\end{figure*}

\subsection{TC-AE}

Based on the above studies, we propose TC-AE, a ViT-based image tokenizer powered by a large and semantically structured token space. An overview of the architecture is shown in Fig.~\ref{fig:main_arch}.

TC-AE consists of a ViT encoder, a latent bottleneck, and a structurally symmetric decoder.
The first innovation lies in the encoder design.
Concretely, the encoder begins with a patch embedding layer using a small patch size $p$ to produce high-resolution image tokens, reducing information loss at the pixel-to-token compression.
These fine-grained tokens are processed by the first $M$ ViT blocks to capture rich visual details and semantic structure.
An intermediate bottleneck then compresses the token sequence to 1/4 length, yielding a compact and structured intermediate representation.
The compressed tokens are further processed by the remaining $N$ ViT blocks, after which a second bottleneck produces the final compact latent representation for downstream generative modeling.

The second innovation in TC-AE lies in its training scheme, which jointly optimizes the tokenizer with a self-supervised objective. This encourages the ViT encoder to learn latent representations with stronger semantic regularization, without requiring external large-scale pretraining for foundation models. Compared to VTP~\citep{vtp}, TC-AE adopts a lightweight training scheme, making it practical to deploy under limited computational resources (see Sec.~\ref{sec:ablaton:dinov2}).

The overall training objective of TC-AE combines the standard reconstruction loss with the self-supervised objective:
\begin{equation}
\mathcal{L}_{\text{TC-AE}} = \alpha \mathcal{L}_{\text{rec}} + \mathcal{L}_{\text{iBOT}}.
\end{equation}

The reconstruction loss $\mathcal{L}_{\text{AE}}$ is defined as:
\begin{equation}
\mathcal{L}_{\text{rec}} = \mathcal{L}_{\text{pix}} + \lambda_p \mathcal{L}_p + \lambda_g \mathcal{L}_g,
\end{equation}
where $\mathcal{L}_{\text{pix}}$ denotes the pixel-level reconstruction loss, implemented as a $\ell_1$ term between original image and reconstructed image.
$\mathcal{L}_p$ is a perceptual loss that captures high-level semantic discrepancies, and $\mathcal{L}_g$ is an adversarial loss that encourages the realism of reconstructed images.

\section{Experiments}
\subsection{Implementation Details}

\noindent \textbf{Tokenizer training.}
All experiments are conducted on ImageNet-1K~\citep{imagenet} with an input resolution of $256 \times 256$.
Compression layers within the ViT encoder are implemented using two-layer convolutional modules.
The final token-to-latent bottleneck is realized via a pixel-shuffle operation followed by an MLP. For ablation studies, the tokenizer is trained for $50$ epochs with a batch size of $512$, using the $\mathcal{L}_{\text{pixel}}$, $\mathcal{L}_p$, and $\mathcal{L}_{\text{iBOT}}$.
For scale-up experiments, we further incorporate the adversarial loss $\mathcal{L}_g$ and finetune the decoder for additional $16$ epochs.
More details and ablations are provided in the appendix.

\noindent \textbf{Evaluation.}
The reconstruction quality is evaluated using SSIM~\citep{ssim}, PSNR~\citep{psnr}, FID~\citep{fid}, and LPIPS~\citep{lpips} on ImageNet-50k validation set.
We adopt LightningDiT from RAE~\citep{rae} to evaluate the generative capability of the tokenizer. The model is trained for $80$ epochs with a batch size of $1024$ for ablation studies. For system-level comparason, the model is trained for $800$ epochs. The generative performance is measured using FID and Inception Score (IS)~\citep{inception-score}.

\begin{figure}[t]
    \centering
    \includegraphics[width=.95\linewidth]{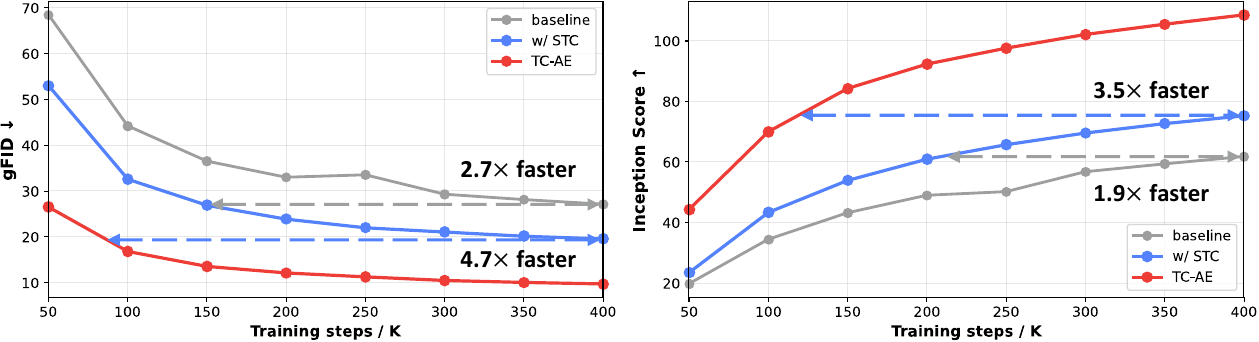}
\caption{Diffusion model convergence comparison. Both staged token compression strategy and joint self-supervised training accelerate the diffusion convergence.}
    \label{fig:exp:convergence}
\end{figure}

\begin{figure*}[t]
    \centering
    \includegraphics[width=0.98\linewidth]{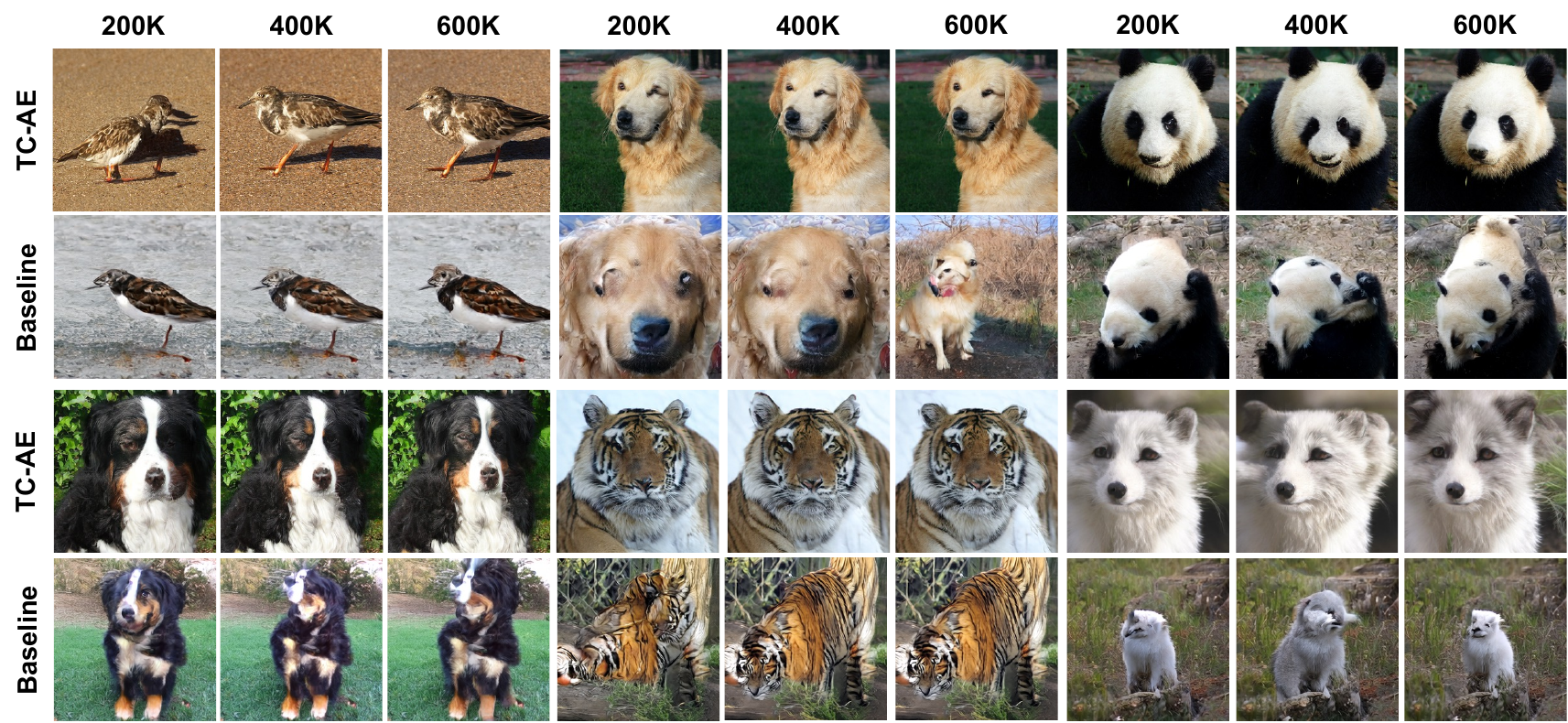}
    \caption{Qualitative comparison between TC-AE and the baseline at $256 \times 256$ resolution.
     Images are sampled from DiT checkpoints at different training steps using the same initial noise. TC-AE produces more visually appealing results compared to the baseline.
    }
    \label{fig:exp:qualitative}
\end{figure*}

\subsection{Synergistic Effects of Token Number Scaling and Structuring}

\noindent \textbf{Accelerating diffusion model convergence.}
We evaluate the convergence behavior of DiT trained on latents produced by different tokenizers under identical training settings.
As shown in Fig.~\ref{fig:exp:convergence}, introducing staged token compression already leads to a noticeable acceleration compared to the baseline: measured by gFID and IS, convergence is accelerated by $2.7\times$ and $1.9\times$, respectively.
When further incorporating self-supervised token structuring, the full TC-AE model achieves substantially faster convergence, reaching the same gFID with $4.7\times$ fewer training iterations and comparable IS with $3.5\times$ fewer steps.

We further provide some qualitative comparison between TC-AE and baseline in Fig.~\ref{fig:exp:qualitative}. 
These results indicate that staged token compression and self-supervised training contribute complementary benefits. By reducing structural information loss during token-to-latent compression and learning a more semantically organized latent space, TC-AE provides a more favorable representation for diffusion model optimization, leading to significantly improved training efficiency.

\begin{table}[tbp]
  \centering
  \captionsetup{justification=centering}
  \caption{Staged token compression improves generation performance.}
  \resizebox{0.75\linewidth}{!}{ 
    \begin{tabular}{ccccccc}
    \toprule
    Staged token compression & rFID$\downarrow$ & PSNR$\uparrow$ & LPIPS$\uparrow$ & SSIM$\uparrow$ & gFID$\downarrow$ & IS$\uparrow$ \\
    \midrule
    \rowcolor{gray!15} \multicolumn{7}{c}{\textit{w/o SSL}} \\
    \midrule
         \xmark & 1.33  & 25.13 & 0.068 & 0.794 & 44.72 & 33.62 \\
         \cmark & 0.75  & 25.72 & 0.072 & 0.807 & 32.92 & 43.58 \\
    \midrule
    \rowcolor{gray!15} \multicolumn{7}{c}{\textit{w/ SSL}} \\
    \midrule
         \xmark & 0.90  & 24.84 & 0.085 & 0.772 & 25.36 & 52.38 \\
         \cmark & 0.90  & 24.21 & 0.098 & 0.749 & 16.39 & 71.33 \\
    \bottomrule
    \end{tabular}}%
  \label{tab:ablation:staged}%
\end{table}%

\noindent \textbf{Effectiveness of staged token compression.}
Table~\ref{tab:ablation:staged} studies the effect of staged token compression, both with and without self-supervision. 

Without SSL, introducing staged token compression consistently improves both reconstruction and generation quality. Specifically, rFID is significantly reduced from $1.33$ to $0.75$, while improving PSNR and SSIM. More importantly, it leads to a substantial improvement in generative performance, reducing gFID from $44.72$ to $32.92$ and increasing IS from $33.62$ to $43.58$. These results indicate that staged token compression alone effectively alleviates the token-to-latent bottleneck and enhances the generatability of the latent representation. When combined with SSL, staged token compression yields even larger gains in generation quality. While reconstruction metrics remain comparable, gFID is further reduced from $25.36$ to $16.39$, and IS increases markedly from $52.38$ to $71.33$. This suggests that staged token compression and semantic token structuring via SSL are complementary: staged token compression preserves semantic information during compression, while SSL further improve the latent structure to benefit downstream generation.

\noindent \textbf{Effectiveness of self-supervision.}
We validate its effectiveness by comparing models trained with and without SSL, and report results under different token numbers by sweeping patch sizes.

As shown in Table~\ref{tab:ablation:ssl}, SSL consistently improves generative performance across all patch sizes, substantially reducing gFID and increasing IS. This confirms that SSL is effective for enhancing the generatability of the latent representation, regardless of token resolution. On the other hand, incorporating SSL generally degrades reconstruction quality, consistent with prior works that directly align latent with pretrained foundation models~\citep{vavae, gigatok}.
We further observe that this effect is more pronounced when the number of tokens is small (e.g., $p=64$), but progressively diminishes as the token 68esolution increases. At smaller patch sizes (e.g., $p=8$), reconstruction performance with SSL becomes comparable to the baseline, while generative performance remains significantly improved.
\begin{table}[tbp]
  \centering
  \captionsetup{justification=centering}
  \caption{Self-supervision improves generative performance across token numbers.}
  \resizebox{0.68\linewidth}{!}{
    \begin{tabular}{c c c c c c c c}
    \toprule
    ViT patch size & SSL & rFID$\downarrow$ & PSNR$\uparrow$ & LPIPS$\uparrow$ & SSIM$\uparrow$ & gFID$\downarrow$ & IS$\uparrow$ \\
    \midrule
    \multirow{2}{*}{64}
      & \xmark &  14.97 & 21.93 & 0.302 & 0.602 & 65.23 & 20.84 \\
      & \cmark &  30.96 & 20.25 & 0.383 & 0.501 & 64.96 & 24.89  \\
    \midrule
    \multirow{2}{*}{32}
      & \xmark & 5.31  & 23.72 & 0.177 & 0.721 & 37.09 & 38.51 \\
      & \cmark & 4.63  & 22.15 & 0.195 & 0.632 & 22.93 & 57.22  \\
    \midrule
    \multirow{2}{*}{16}
      & \xmark & 1.12  & 25.03 & 0.080  & 0.779 & 31.37 & 45.49  \\
      & \cmark & 1.27  & 23.76 & 0.104 & 0.723 & 17.22 & 69.00  \\
    \midrule
    \multirow{2}{*}{8}
      & \xmark & 1.33  & 25.13 & 0.068 & 0.794 & 44.72 & 33.62 \\
      & \cmark & 0.90   & 24.84 & 0.085 & 0.772 & 25.36 & 52.38 \\
    \bottomrule
    \end{tabular}}
  \label{tab:ablation:ssl}
\end{table}

\subsection{Synergistic Effects of Token Number Scaling and Parameter Scaling}

\begin{wrapfigure}[15]{r}{0.5\linewidth}
    \centering
    \vspace{-3mm}
    \includegraphics[width=\linewidth]{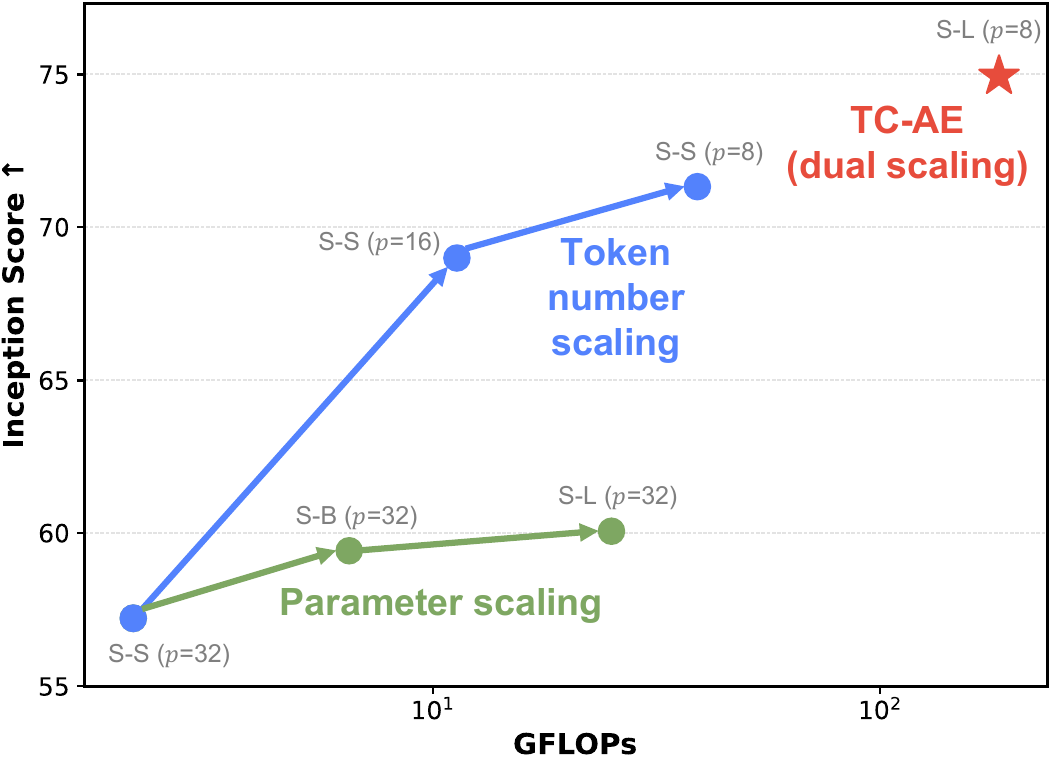}
    \caption{Synergy of token number scaling and parameter scaling.}
    \label{fig:exp:scaling}
\end{wrapfigure}

Prior work~\citep{vitok} observes that increasing decoder parameters can improve the generatability of the latent representation. To compare the effect between model capacity and token capacity, we vary the decoder size across S-\{S, B, L\} and sweep the patch size $p \in \{32, 16, 8\}$ to control the number of image tokens. For each configuration, we train the downstream DiT model under identical settings.

As shown in Fig.~\ref{fig:exp:scaling}, increasing the number of tokens or enlarging the decoder consistently improves generative performance. However, for comparable computational cost (GFLOPs), token number scaling yields stronger improvements than parameter scaling.
More importantly, the two scaling dimensions are complementary: combining larger token spaces with higher decoder capacity leads to further gains, demonstrating a synergistic effect between token number and model size.

\subsection{System-Level Comparison}

We compare TC-AE with existing tokenizers and downstream generative models on ImageNet 256×256 in Table~\ref{tab:exp:main}.

Among deep compression tokenizers operating with only 64 tokens, TC-AE significantly outperforms prior methods in generative quality.
When paired with the same DiT backbone, TC-AE reduces gFID from 26.44 (DC-AE) and 17.31 (DC-AE-1.5) to 7.16 without classifier-free guidance (CFG), and further to 2.57 with CFG enabled.
Notably, this improvement is achieved with substantially lower computational cost: TC-AE requires only about 164 GFLOPs, compared to 607 GFLOPs for DC-AE.
This substantial margin demonstrates that TC-AE effectively mitigates representation collapse and preserves generative structure under deep compression, while being significantly more computationally efficient.Compared with low-compression tokenizers, TC-AE achieves strong generative performance. For instance, TC-AE with only 64 tokens matches or outperforms methods using 256 tokens such as ViTok (2.45) and VAR (2.95), while also exceeding several billion-parameter generator setups in LlamaGen.

Overall, these results demonstrate that TC-AE bridges the performance gap between deep compression and low-compression tokenizers, enabling high-quality image generation while retaining the efficiency benefits of deep compression.

\begin{table*}[t]
  \centering
\caption{System-level comparison of tokenizers and downstream generative models on ImageNet 256$\times$256.
The CFG scale is set to 1.4 for TC-AE. $\dagger$: Results are taken from the original paper; additional comparisons are provided in the appendix.}
  \resizebox{\linewidth}{!}{ 
    \begin{tabular}{ccccccc}
    \toprule
    \textbf{Tokenizer} & \textbf{Latent Tokens} & \textbf{rFID $\downarrow$} & \textbf{Generative Models} & \textbf{Param} & \textbf{gFID (w/o CFG) $\downarrow$} & \textbf{gFID (w/ CFG) $\downarrow$} \\
    \midrule
    \rowcolor{gray!15} \multicolumn{7}{c}{\textit{low compression tokenizer}} \\
    \midrule
    MaskGIT~\citep{maskgit} & 256   & 2.28  & MaskGIT~\citep{maskgit} & 227M  & 6.18  & - \\
    VQGAN~\citep{vqgan} & 256   & 0.59  & LlamaGen~\citep{llamagen} & 3B    & 9.38  & 2.18 \\
    VAR-Tok.~\citep{var} & 680   &   1.00    & VAR-d20~\citep{var} & 600M  & -     & 2.95 \\
    LlamaGen-Tok.~\citep{llamagen} & 256   & 2.19  & LlamaGen~\citep{llamagen} & 1.4B  & -     & 3.09 \\
    VAE~\citep{ldm}   & 256   & 0.53  & MAR~\citep{mar}   & 943M  & 2.35  & 1.55 \\
    VAE~\citep{ldm}   & 4096  & 0.27  & LDM-4~\citep{ldm} & 400M  & -     & 3.60 \\
    SD-VAE~\citep{sdvae} & 256   &   0.61    & DiT~\citep{dit}   & 600M  & 9.62  & 2.27 \\
    ViTok S-B~\citep{vitok} & 256   & 0.18  & DiT~\citep{dit}   & 675M  & -     & 2.45 \\
    MAETok~\citep{maetok} & 256   &   0.48    & LightningDiT~\citep{vavae} & 675M  & 2.21  & 1.73 \\
    GigaTok-B-L~\citep{gigatok} & 256   &    0.51   & LlamaGen~\citep{llamagen} & 111M  & -     & 3.33 \\
    \midrule
    \rowcolor{gray!15} \multicolumn{7}{c}{\textit{deep compression tokenizer}} \\
    \midrule
    DC-AE-1.5$^\dag$~\citep{dc-ae-1.5} & 64    & 0.26  & DiT~\citep{dit}   & 675M  & 17.31 & - \\
    DC-AE$^\dag$~\citep{dc-ae} & 64    & 0.26  & DiT~\citep{dit}   & 675M  & 26.44 & - \\
    \cellcolor{rowblue}{TC-AE} & \cellcolor{rowblue}{64} & \cellcolor{rowblue}{0.35} & \cellcolor{rowblue}{DiT~\citep{rae}} & \cellcolor{rowblue}{675M} & \cellcolor{rowblue}{7.16} & \cellcolor{rowblue}{2.57} \\
    \bottomrule
    \end{tabular}}%
  \label{tab:exp:main}%
\end{table*}%

\begin{table*}[t]
  \centering
  \captionsetup{justification=centering}
  \begin{minipage}[c]{0.48\textwidth}
    \scriptsize
    \caption{Layer depth for staged token compression.}
    \centering
    \begin{tabular}{ccccc}
    \toprule
    Layer depth ($M$) & rFID$\downarrow$  & SSIM$\uparrow$  & IS$\uparrow$ (w/o cfg) & IS$\uparrow$ (w/ cfg) \\
    \midrule
    0     & 1.27  & 0.723 & 69.00    & 312.45 \\
    3     & 0.93  & 0.728 & 66.68    & 311.45 \\
    \rowcolor{rowblue}
    6     & 0.90  & 0.749 & 71.33    & 326.47 \\
    9     & 0.92  & 0.756 & 62.61    & 304.26 \\
    12    & 0.90  & 0.772 & 52.38    & 257.93 \\
    \bottomrule
    \end{tabular}
    \label{tab:stage depth}
  \end{minipage}
  \hfill
  \begin{minipage}[c]{0.48\textwidth}
    \scriptsize
    \caption{Training with other self-supervised learning methods.}
    \centering
    \begin{tabular}{ccccccc}
    \toprule
    SSL method & rFID$\downarrow$ & PSNR$\uparrow$ & LPIPS$\uparrow$ & SSIM$\uparrow$ & gFID$\downarrow$ & IS$\uparrow$ \\
    \midrule
    DINO   & 1.34 & 23.94 & 0.101 & 0.732 & 20.38 & 62.12 \\
    DINOv2 & 1.34 & 24.48 & 0.103 & 0.759 & 28.22 & 49.19 \\
    \rowcolor{rowblue}
    iBOT   & 1.27 & 23.76 & 0.104 & 0.723 & 17.22 & 69.00 \\
    \bottomrule
    \end{tabular}
    \label{tab:ablation:other-ssl}
  \end{minipage}
\end{table*}

\subsection{Design Space of TC-AE}
\noindent \textbf{Layer depth for each stage.}
We study where to apply staged token compression in the ViT encoder by varying the split between the two stages. The encoder contains $12$ Transformer layers.

As shown in Table~\ref{tab:stage depth}, reconstruction quality (SSIM) consistently improves as $M$ increases. This trend indicates that allocating more layers to process high-resolution tokens benefits reconstruction by preserving finer visual details. In contrast, TC-AE achieves the best generative performance $M=6$. We hypothesize that this behavior reflects a trade-off between semantic structure learning and token-to-latent compression.
The first stage primarily learns rich semantic structure from the large token space, while the second stage, together with the final bottleneck, acts as a compression buffer between high-resolution tokens and the compact latent. Increasing $M$ strengthens semantic representation learning but reduces the capacity available for compression, which may amplify structural information loss during token-to-latent transformation. At $M=6$, this trade-off reaches an optimal balance, resulting in the strongest generative performance. Based on these observations, we adopt $M=6$ as the default configuration in the scale-up experiment.

\noindent \textbf{Joint training with different SSL methods.}
\label{sec:ablaton:dinov2}
We further compare different self-supervised objectives for joint tokenizer training.
In addition to iBOT, we evaluate other representative self-distillation methods, including DINO~\citep{dino} and DINOv2~\citep{dinov2}, where DINOv2 is also adopted in the concurrent work VTP~\citep{vtp}.

As shown in Table~\ref{tab:ablation:other-ssl}, iBOT consistently achieves stronger generative performance than DINO, while maintaining comparable reconstruction quality.
Furthermore, joint training with DINOv2 leads to notably weaker gains under our training setting with a batch size of $512$ on $8$ NVIDIA GPUs, indicating limited effectiveness in the moderate-batch regime.
Overall, these results suggest that iBOT offers a lightweight, stable, and practical self-supervised objective for joint tokenizer training under realistic computational budgets.

\section{Conclusion}

In this work, we study deep compression autoencoders from the perspective of the token space. We show that naively scaling the number of image tokens fails to benefit generation due to severe structural information loss at the compression bottleneck. To address this issue, we propose to redistribute token-to-latent compression across encoder stages, enabling effective token number scaling for generation. In addition, we introduce self-supervised objective to further structure the token space, leading to more generative-friendly latent representations. Extensive experiments on ImageNet demonstrate that TC-AE significantly improves both reconstruction and generative performance under deep compression, accelerates diffusion model convergence, and achieves a favorable efficiency–quality trade-off compared to existing tokenizers.

\section{Acknowledgement}
This work was supported by Ant Group Research Intern Program.

\newpage

\bibliographystyle{assets/plainnat}
\bibliography{ref}

\clearpage
\appendix

\section{TC-AE Implementation Details}
The detailed training configurations of TC-AE are provided in Table~\ref{tab:exp:detail}.

\noindent \textbf{Self-supervision training.}
We adopt the same data augmentation pipelines as in the corresponding self-supervised learning methods~\citep{dino, ibot, dinov2}. To ensure stable joint optimization with the reconstruction objective, we reduce the learning rate.

\noindent \textbf{Adversarial training.}
We adopt the adversarial training setup from RAE~\citep{rae}. The discriminator is built upon a frozen DINO-S/8 backbone~\citep{dino}, which has been shown to stabilize training and prevent the decoder from exploiting adversarial patch-level artifacts. All inputs are resized to $224 \times 224$ before being fed into the discriminator.

\begin{table}[h]
  \centering
  \captionsetup{justification=centering}
  \caption{TC-AE training details.}
    \resizebox{0.7\linewidth}{!}{ 
    \begin{tabular}{ccc}
    \toprule
       \textbf{Component}   &  \textbf{Joing Training} &  \textbf{Decoder Finetuning} \\
    \midrule
    Training data & ImageNet & ImageNet \\
    Training epochs & 50    & 16 \\
    Batch size & 512   & 256 \\
    Learning Rate schedule & Cosine Decay & Cosine Decay \\
    Optimizer betas & (0.9, 0.95) & (0.9, 0.95) \\
    Weight decay & 0.04  & 0 \\
    Base learning rate & 1e-3  & 2e-4 \\
    Minimum learning rate & 1e-6  & 2e-5 \\
    Warmup epoch & 5     & 1 \\
    Optimizer & AdamW & AdamW \\
    $\alpha$ & 0.1     & 0 \\
    $\lambda_p$ & 1    & 1 \\
    $\lambda_g$ & 0     & 0.75 \\
    Discriminator start epoch & -     & 6 \\
    Adversarial loss start epoch & -     & 8 \\
    \bottomrule
    \end{tabular}}%
  \label{tab:exp:detail}%
\end{table}%

\section{More Ablations and Results}

\noindent \textbf{Compression module design in ViT.}
We compare two alternative designs for the token compression module within the ViT encoder: a pixel-shuffle followed by an MLP, and a lightweight two-layer convolution.
All experiments are conducted with a patch size of $p=8$.
For the pixel-shuffle-based design, token compression is implemented by first merging each $2\times2$ group of neighboring tokens into a single token via pixel shuffle, followed by an MLP that downsamples the channel dimension by a factor of four, resulting in a $4\times$ reduction in token count.

\begin{table}[h]
  \centering
  \captionsetup{justification=centering}
  \caption{Compression module design in ViT.}
    \resizebox{0.75\linewidth}{!}{ 
    \begin{tabular}{ccccccc}
    \toprule
    Compression module & rFID$\downarrow$ & PSNR$\uparrow$ & LPIPS$\uparrow$ & SSIM$\uparrow$ & gFID$\downarrow$ & IS$\uparrow$ \\
    \midrule
    Pixel shuffle + MLP & 0.89  & 25.72 & 0.070  & 0.808 & 37.07 & 38.91 \\
\cellcolor{rowblue}{Two-layer convolution} &
\cellcolor{rowblue}{0.75} &
\cellcolor{rowblue}{25.72} &
\cellcolor{rowblue}{0.072} &
\cellcolor{rowblue}{0.807} &
\cellcolor{rowblue}{32.92} &
\cellcolor{rowblue}{43.58} \\
    \bottomrule
    \end{tabular}}%
  \label{tab:exp:compress_design}%
\end{table}%

As shown in Table~\ref{tab:exp:compress_design}, the convolutional module consistently yields better generative performance in terms of gFID and IS. We conjecture that this improvement stems from aggregating local neighborhood information, as convolution explicitly models spatial locality. Based on these observations, we adopt the two-layer convolution as the default compression module in TC-AE.

\noindent \textbf{Weight of $\alpha$.}
We examine the effect of the weighting factor $\alpha$, which controls the trade-off between reconstruction and SSL objectives during tokenizer training.

\begin{table}[h]
  \centering
  \captionsetup{justification=centering}
  \caption{Trainging weights between reconstruction and self-supervision.}
    \resizebox{0.55\linewidth}{!}{ 
    \begin{tabular}{ccccccc}
    \toprule
    $\alpha$ & rFID$\downarrow$ & PSNR$\uparrow$ & LPIPS$\uparrow$ & SSIM$\uparrow$ & gFID$\downarrow$ & IS$\uparrow$ \\
    \midrule
    \rowcolor{rowblue} 0.1   & 1.27  & 23.76 & 0.104 & 0.723 & 17.22 & 69.00 \\
    1.0   &    1.31  & 24.79 & 0.092 & 0.768 & 27.74 & 48.92 \\
    10    &  1.26  & 25.22 & 0.086 & 0.789 & 34.10  & 40.62  \\
    \bottomrule
    \end{tabular}%
  \label{tab:exp:alpha}}%
\end{table}%

As shown in Table~\ref{tab:exp:alpha}, increasing $\alpha$ improves reconstruction at the expense of generative quality.
This reflects a trade-off between emphasizing high-frequency details and learning semantically structured latents.
We therefore set $\alpha=0.1$ as the default in all experiments.

\noindent \textbf{Expanding latent channel number to 256.}
Table~\ref{tab:exp:more_channel} further examines the effect of increasing the latent channel dimension from $c=128$ to $c=256$.
Overall, both staged token compression (STC) and self-supervised learning (SSL) consistently improve reconstruction and generative performance relative to their corresponding baselines when the channel dimension is expanded.

Despite these gains, models with $c=256$ exhibit inferior generative quality compared to their $c=128$ counterparts, as indicated by higher gFID and lower IS.
This observation suggests that simply increasing the latent channel dimension may aggravate representation collapse.
In practice, $c=128$ provides a more favorable balance between reconstruction fidelity and latent generatability.

\begin{table}[h]
  \centering
  \captionsetup{justification=centering}
  \caption{Expanding latent channel number to 256.}
    \resizebox{0.7\linewidth}{!}{ 
    \begin{tabular}{ccccccccc}
    \toprule
    Channel & SSL   & STC   & rFID$\downarrow$ & PSNR$\uparrow$ & LPIPS$\uparrow$ & SSIM$\uparrow$ & gFID$\downarrow$ & IS$\uparrow$  \\
    \midrule
    \multirow{3}[2]{*}{128} &       &       & 1.33  & 25.13 & 0.068 & 0.794 & 44.72 & 33.62 \\
          &       & \cmark     & 0.75  & 25.72 & 0.072 & 0.807 & 32.92 & 43.58 \\
          & \cmark     &       & 0.57  & 27.98 & 0.037 & 0.875 & 39.93 & 36.49 \\
    \midrule
    \multirow{3}[2]{*}{256} &       &       & 0.73  & 28.06 & 0.040  & 0.879 & 47.33 & 30.73 \\
          &       & \cmark     & 0.57  & 27.98 & 0.037 & 0.875 & 39.93 & 36.49 \\
          & \cmark   &   &  1.23  & 24.99 & 0.091 & 0.780  & 24.27 & 55.20  \\
    \bottomrule
    \end{tabular}}%
  \label{tab:exp:more_channel}%
\end{table}%

\noindent \textbf{Additional comparison with DC-AE.}
We download the official checkpoint\footnote{\url{https://huggingface.co/mit-han-lab/dc-ae-f32c32-in-1.0-256px}} from the authors of DC-AE~\citep{dc-ae} and use it as the tokenizer.
We then train diffusion models using DC-AE and TC-AE under same configurations. The latent shape of TC-AE is kept the same as DC-AE for a fair comparison. The results are reported in Table~\ref{tab:exp:dcae}.
TC-AE consistently achieves better generative performance than DC-AE.
These results further demonstrate that the proposed staged token compression and joint self-supervised training effectively improve latent generatability.

\begin{table}[h]
  \centering
  \captionsetup{justification=centering}
  \caption{Additional comparison with DC-AE. The CFG scale is set to 1.4.}
    \resizebox{0.8\linewidth}{!}{ 
    \begin{tabular}{cccccc}
    \toprule
    Tokenizer & Latent shape & GFLOPS & rFID$\downarrow$  & gFID (w/o CFG) $\downarrow$ & gFID (w/ CFG) $\downarrow$ \\
    \midrule
    DC-AE & $8\times8\times32$ & 607   & 0.71  & 15.90 & 5.59 \\
    TC-AE & $8\times8\times32$ & 164   & 0.81  & 10.72 & 4.77 \\
    \bottomrule
    \end{tabular}}%
  \label{tab:exp:dcae}%
\end{table}%

\begin{figure*}[h]
    \centering
    \includegraphics[width=0.75\linewidth]{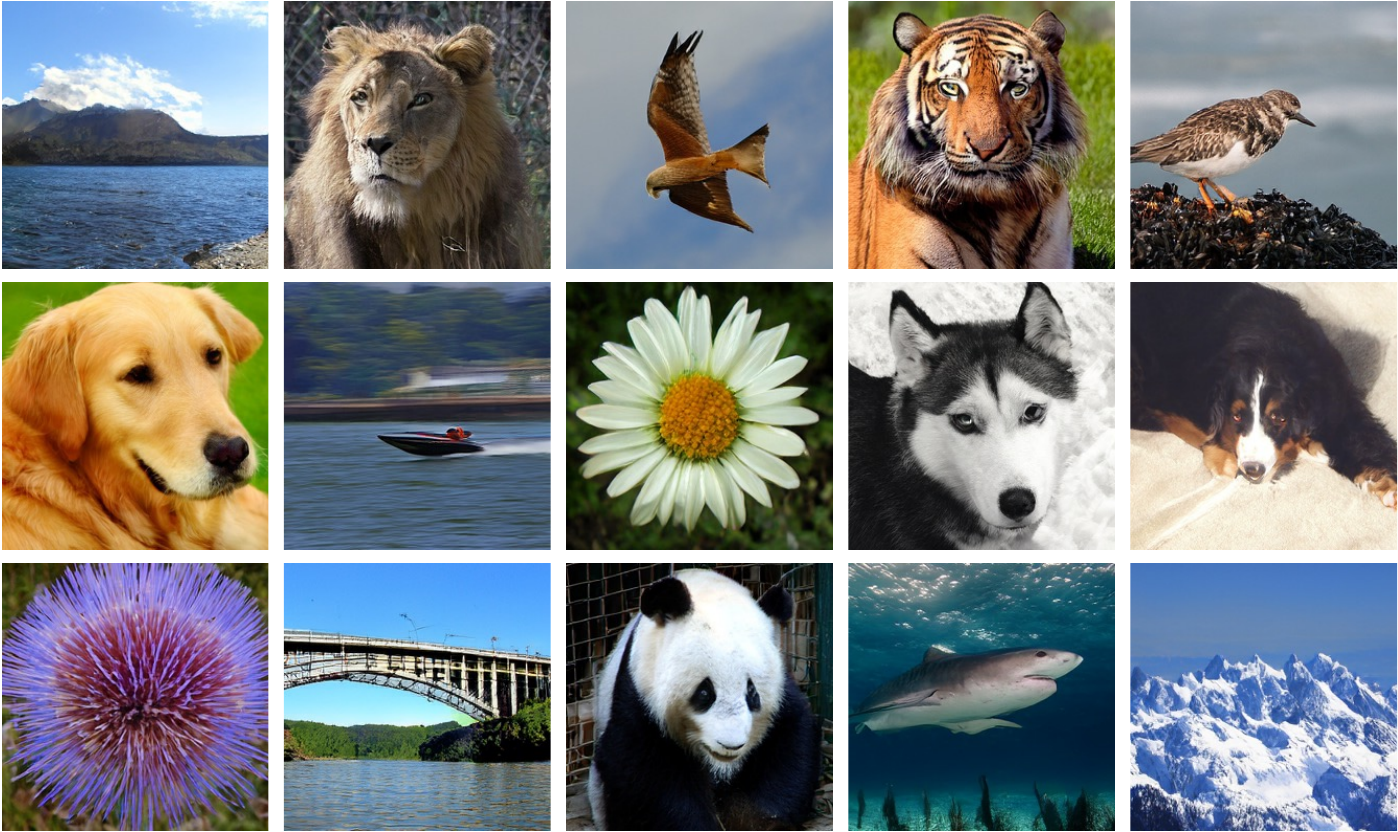}
    \captionsetup{justification=centering}
    \caption{Image generated by diffusion models using TC-AE at $256 \times 256$ resolution.
    }
    \label{fig:exp:qualitative_tcae}
\end{figure*}

\noindent \textbf{Additional visualization.} We provide more generation results in Fig.~\ref{fig:exp:qualitative_tcae}.

\end{document}